# Micro-Attention for Micro-Expression Recognition

**Chongyang Wang[1][†], Min Peng[2][†][*], Tao Bi[1], Tong Chen[3][4]**

[1]UCL interaction centre, University College London, London, United Kingdom

[2]Intelligent Security Center, Chongqing Institute of Green and Intelligent Technology, CAS, Chongqing, China

[3]College of Electronic and Information Engineering, Southwest University, Chongqing, China

[4]Institute of Psychology, CAS, Beijing, China

[†]Equal contribution

**\* Correspondence:**
Min Peng
pengmin@cigit.ac.cn

**Abstract**

Micro-expression, for its high objectivity in emotion detection, has emerged to be a promising modality in affective computing. Recently, deep learning methods have been successfully introduced into the micro-expression recognition area. Whilst the higher recognition accuracy achieved, substantial challenges in micro-expression recognition remain. The existence of micro expression in small-local areas on face and limited size of available databases still constrain the recognition accuracy on such emotional facial behavior. In this work, to tackle such challenges, we propose a novel attention mechanism called micro-attention cooperating with residual network. Micro-attention enables the network to learn to focus on facial areas of interest covering different action units. Moreover, coping with small datasets, the micro-attention is designed without adding noticeable parameters while a simple yet efficient transfer learning approach is together utilized to alleviate the overfitting risk. With extensive experimental evaluations on three benchmarks (CASMEII, SAMM and SMIC) and post-hoc feature visualizations, we demonstrate the effectiveness of the proposed micro-attention and push the boundary of automatic recognition of micro-expression.

## 1    Introduction

Micro-expression is a rapid and subtle facial movement, which can reveal underlying genuine emotions. Typically, people express their emotions consciously by macro-expressions that last from 1/2 to 4 seconds [1] and are easily perceived by humans. Meanwhile, many researches have been focused on enabling computer to learn human emotion by recognizing macro-expressions (for surveys, see [2] [3]). However, psychological studies [4] [5] suggest that macro-expression could be misleading on human emotion recognition. Unlike macro-expression, micro-expression is mostly expressed unconsciously where the genuine emotion can be revealed. For such objectivity, the recognition of micro-expression powered many applications in diverse areas like affect monitoring [4], criminal detection [6], and homeland security [7].

---

\*Code available at https://github.com/CodeShareBot/Micro-Attention-for-Micro-Expression

Previous research on automatic recognition of micro-expression is benefited from successful feature-engineering works on macro-expression. First, Zhao et al. [8] applied Local Binary Pattern on Three Orthogonal Planes (LBP-TOP) to extract facial features from image sequences containing micro-expression and used Support Vector Machine (SVM) to classify the features. Later, Wang et al. [9] further designed LBP-Six-Intersection-Points (LBP-SIP) for the feature extraction. After that, Li et al. [10] compared LBP-TOP, Histogram of Oriented Gradients (HOG) and Histogram of Image Gradient Orientation (HIGO) for the feature extraction of micro-expression videos. On the other hand, a study using tensor engineering has been seen in [46]. Except for these evolved from analyses on facial macro-expression, other features proposed in video analysis have also inspired the micro-expression research. Recently, Optical Flow [11] has been used to develop the feature designing for micro-expression. Liu et al. [12] proposed a method called Main Directional Mean Optical-flow (MDMO) to capture the subtle facial movement for micro-expression recognition. Later, Xu et al. [13], based on Optical Flow, designed a fine-grained sequence alignment method and an optical flow direction optimization strategy for the recognition of micro-expression. Aside from the recognition solely based on micro-expression data, a study by Zhu et al [47] proposed to transfer the knowledge of larger affective speech data to the recognition of micro-expression. Generally, methods based on feature engineering that require dependent data pre-processing are more suitable for off-line analysis, while for faster and even real-time analysis of micro-expression, better end-to-end methods are needed.

On the other hand, as micro-expression is demonstrated through the dynamic facial movement, most existing works have focused on the recognition from videos. However, a large percentage of frames within a video clip could be redundant for the recognition because the micro-expression is transient and only exists in a few frames. A recent research [14] proposed a comprehensive pipeline for automatic apex spotting as well as proved the merit of using apex frame within a video clip for micro-expression recognition, while the accuracy achieved was higher than traditional methods that used full video clips as input.

In this paper, we follow the idea of deep learning in designing end-to-end network for micro-expression recognition. To lessen over-fitting, we perform transfer learning to aid the training on micro-expression databases, which has proved to be efficient in applying deep neural network on small databases in a recent work [17]. Furthermore, the recognition is performed on apex frames within each video sequence, which aimed to get rid of redundant information as well as directly take the advantage of larger macro-expression datasets that mostly comprising images. In different with the study [17], a novel attention mechanism called micro-attention is designed to help in focusing on the region of interest, considering that the micro-expression is expressed by a fleeting movement of local areas on face. To the best of our knowledge, this paper is the first to apply attention mechanism on the recognition of micro-expression. Technically, we modified the way of integrating residual network with attention mechanism in contrast to a previous work [29], with an aim of reducing the number of parameters. Namely, we compute the attention map using the self-outputs from each residual block at different scales. Through experiments, we showed that our architecture is able to achieve better recognition accuracy with smaller parameter size. Extra feature visualizations are also imaging the effectiveness of micro-attention on capturing the facial areas of interest. The main contributions are summarized as below:

- A novel micro-attention design for micro-expression recognition is proposed without introducing notable extra parameters. Cooperating with residual network, such attention design also takes the advantage of multi-scale spatial features and enables the network to focus on the area where micro-expression exactly happens thus improve the recognition accuracy. Our method is expected to be useful for other computer vision tasks, especially when the region-of-interest only exists within local areas and the size of database is small.

- Extensive evaluations on three benchmark databases demonstrate the effectiveness of the proposed attention unit comparing with other state-of-the-art methods. Particularly, through post-hoc feature visualizations, the power of micro-attention in learning spatial attentions that accord with respective facial action units (AUs) is illustrated. Meanwhile, the potential for other facial expression tasks reveals.

## 2    Related Research Works

The deep learning approach, as a group of machine learning especially neural network methods, has produced many successful case studies in image processing, video analysis and speech recognition. Without hand-crafted feature engineering, an end-to-end neural network model is able to classify and predict by learning from large sets of high dimensional (and low-level representation of) data [15]. Convolutional neural network (CNN), as one of the most widely applied deep learning approaches, is currently the leading method in many image-related areas, like large-scale object recognition [18] and face recognition [19]. First introduced in study [20], the CNN has been modified a lot within the past years in a layers-increasing and block-designing manner, popular successors of which are AlexNet [21], VGG-Net [22] and GoogLeNet [23]. Despite the differences in network architecture, deep learning models are benefited from the ability of learning high dimensional representation from large datasets.

Recently, we have seen several studies that applied deep learning for the micro-expression recognition. At first, researchers [43][44] used CNNs to extract features from micro-expression videos and further applied the classifier like SVM to acquire the classification results. Comparing with traditional hand-crafted features, these deep features showed better performance on characterizing micro-expressions. Then, Peng et al. [16] designed the first end-to-end middle-size neural network called Dual Temporal Scale Convolutional Neural Network (DTSCNN) for micro-expression recognition. The DTSCNN has two temporal channels designed for data that share different temporal nature, e.g. cameras used for data collection has different framerates. To partially avoid over-fitting, each channel only has 4 convolutional layers and 4 pooling layers. The recognition rate achieved is around 10% higher than some previous state-of-the-art methods (e.g. MDMO, FDM). More recently, Huai-Qian et al. [32] proposed to train a network with convolutional layers and recurrent layers for micro-expression recognition. Instead of using data augmentation for the dataset, they extracted optical flow features to enrich the input at each timestep or with a given temporal length. However, the results achieved are only around chance level which could be due to the practice of using deep network on small datasets. Additionally, for the nature of micro-expression that the duration is less than 1/2 second, it may be inappropriate to train the network on the full video clips. A more recent study [17] proposed to use a pre-trained residual network for micro-expression recognition with apex frames within each video sequences at the Facial Micro-Expression Grand Challenge (MEGC 2018) [41] in the 13th IEEE Conference on Automatic Face and Gesture Recognition and won first place in all tasks. This work also proved the advantage of using apex frames instead of full video sequences for micro-expression recognition with deep learning method. So far, none of works have considered that micro-expression is only expressed within small-local facial areas. Additionally, the size of available micro-expression datasets is much smaller than that of traditional database fed into CNN, which could cause serious over-fitting problem.

In computer vision domain, the attention mechanism [25] is proposed to find the region of interest

(ROI) within an image for respective tasks and highlight the representation of the location. While for micro-expression recognition, such mechanism may help the network to focus on the ROI on the face and reduce the negative influence from irrelevant facial areas and background. A successful application of attention mechanism for micro-gesture recognition is seen in [29], which designed an attention block and added it to the end of the last convolutional layer within each residual block. The spatial attention of micro-gesture was learned and lead to an improvement in recognition accuracy. However, the attention block proposed in that study create notable extra parameters for training, which further increased the training time and complexity.

In this work, we designed a simplified yet efficient attention mechanism to learn the spatial focus of micro-expression within each image. Specially, the proposed trainable micro-attention unit is designed without increasing notable parameters for training and would be discussed in detail in next section. On the other hand, the transfer learning (for a survey, see studies in [30][31]) has proved to be very useful for using the knowledge in source domain to help the learning in target domain, especially when the size of target dataset is too small to train a network. When the source and target domain share a similar data structure, a simple yet efficient implementation of transfer learning is to initialize the network on larger-relevant databases and then fine tune it with the target database. We would follow this idea to take the advantage of larger macro-expression databases to improve the learning on micro-expression databases in this paper.

## 3    Methodology

In this section, we discuss the proposed methods for micro-expression recognition. First, a residual network is used as a basic architecture. Second, within each residual block, a novel micro-attention unit is integrated to enable the network to focus on the facial areas exhibiting micro-expression. Finally, to train the network, a transfer learning approach is utilized to lessen the over-fitting risk.

### 3.1    Residual Network

The residual network [24] previously achieved a state-of-the-art performance on image-base recognition tasks without increasing network complexity. The residual network usually uses a stack of residual blocks to build the network, while a classic residual block is shown in Figure 1. Within a residual block, a shortcut connection would be added to perform identity mapping (achieved by element-wise summarization), which helps to reduce the degradation problem. The shortcut connection also accelerates the training process and nearly introduces no extra parameter and computation. In this paper, we designed our network with a stack of 10 residual blocks, within each a micro-attention unit is added to learn the spatial attention map. Specially, we call such block as concise and trainable residual attention block. Through training, such network is able to focus on micro-expression on the face.

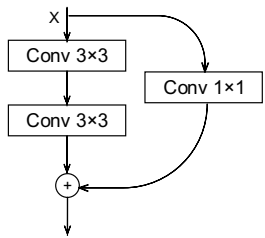

Figure 1. A classic residual block with a shortcut connection that designed for the input and output of the block sharing different dimensions.

## 3.2 Micro-Attention Unit

Three aspects have been considered in designing the residual attention block: 1) the attention unit should be trainable, 2) the attention unit should be compact without increasing notable parameters, 3) the residual scheme should be combined in learning the attention feature.

The proposed residual attention block is shown in Figure 2. Specifically, the proposed trainable micro-attention unit is shown in the dashed-line area, where the average spatial feature map $M(X)$ is computed. Unlike the state-of-the-art attention units that would bring noticeable computation load [26] [29], we compute the attention map using the multi-scale features that self-learned within the residual architecture, which is more straightforward and parameter-saving. Specifically, given an input $X \in \mathbb{R}^{C \times H \times W}$ ($C, H, W$ are the number of channels, height and width respectively) to this block, the three convolutional layers, namely the Conv $1 \times 1$, first Conv $3 \times 3$ and second Conv $3 \times 3$ can produce three feature matrices, namely $\{\mathcal{L}^{c_1}, \mathcal{L}^{c_2}, \mathcal{L}^{c_3}\}$, under three convolutional scales of $1 \times 1$, $3 \times 3$ and $5 \times 5$ respectively. Based on these features of multiple channels, in the micro-attention unit, a channel-wise concatenation is used to compute $\mathcal{L}^{c'} \in \mathbb{R}^{(c_1+c_2+c_3) \times H \times W}$. Then, an embedding implemented with $1 \times 1$ convolution together with channel-wise averaging are performed to produce the average feature map as

$$M(X) = \frac{1}{c'} \sum_{dim=0,i=1}^{c'} \mathcal{L}^{c'} W_* \tag{1}$$

where $W_*$ is a weight matrix of the $1 \times 1$ convolution layer to be learned, $c' = c_1 + c_2 + c_3$ is the number of channels of $\mathcal{L}^{c'}$. Here need to mention that the size of the attentional feature map computed at each residual block is equal with its residual output. After an element-wise multiplication with the current residual output $T(X) = \mathcal{L}^{c_1} + \mathcal{L}^{c_3}$, the output $O(X)$ of the whole residual attention block is computed as

$$O(X) = T(X) \cdot \begin{bmatrix} 1 \\ M(X) \end{bmatrix} \tag{2}$$

Comparing with the output $T(X)$ from the original residual block, the proposed residual attention block only brings in the $T(X) * M(X)$ section which denotes the attention computation. The attention map $M(X)$ would approximately approach zero when the attentional areas are not obviously learned for original output $T(X)$.

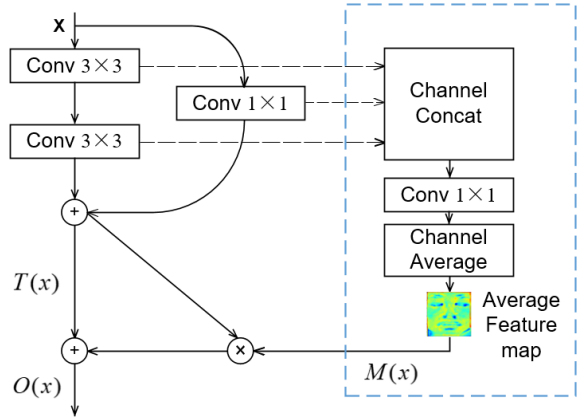

Figure 2. The residual attention block, where the proposed micro-attention unit is shown in dotted-line area.

## 3.3 Transfer Learning

To lessen over-fitting when training deep learning network on micro-expression datasets, the transfer learning approach is considered. Similar with the ideas of transferring the information from macro-expressions to help the recognition on micro-expression [48][49], we implement our transfer learning strategy according to a recent work that also used deep learning in this topic [17]. The proposed steps are shown in Figure 3. First, the original residual network (without micro-attention units) is initialized with the ImageNet database [38]. Then, for the differences between object recognition (ImageNet) and facial expression recognition, such network is further pre-trained on several popular macro-expression databases. The selected macro-expression databases are Cohn-kanade dataset (CK+) [33], Oulu-CASIA NIR&VIS facial expression [34], Jaffe [35], and MUGFE [36]. Details about these databases are provided in next section. Finally, the residual network together with micro-attention units is fine-tuned with micro-expression databases, namely the CASMEII [27], SAMM [28] and SMIC [45].

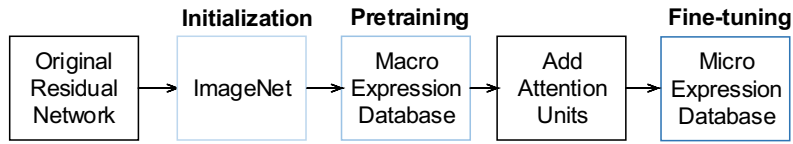

Figure 3. The applied transfer learning process.

## 4 Experiment and Discussion

In this section, we evaluate the proposed method with several popular macro-expression databases (for transfer learning) and three benchmark micro-expression databases. The database description and pre-processing details are first provided. Then, the implementation details are presented. Finally, the analysis on experimental results in comparison with several related works is reported.

### 4.1 Data Preparation

Three micro-expression databases are used in this work, namely the CASMEII [27], SAMM [28] and SMIC [45]. A summary of these three databases is given in Table 1. To avoid the category bias between databases, we apply two strategies to regroup the data in each dataset respectively: i) when CASME II and SAMM are used together for cross-database validations, video clips in each database are regrouped into 5 emotion types based on the intensity distribution of AUs [38] according to the study [39]. The 5 emotions are happiness, surprise, anger, disgust and sadness, while data originally labelled as fear and others are not used; ii) when the three databases are used separately, data in CASMEII and SAMM are also regrouped into the 5 categories as stated while the category of data in SMIC stay unchanged.

The CASMEII and SAMM databases provided the labelling for the onset, apex and offset frames of each video clip. For SMIC, that the apex frame is not marked, we use the frame at the middle of each video clip as the estimated apex frame. Given an apex frame, AAM [40] is used to automatically locate and segment the facial area, and the processed images would be further normalized into 224 × 224 pixels.

To help the training on the three micro-expression databases, a transfer learning approach is employed. During the pre-training step, 4 popular macro-expression databases are used, namely the CK+ [33], Oulu-CASIA NIR&VIS facial expression [34], Jaffe [35], and MUGFE [36].

Table 1 | Summary of the three micro-expression databases (numbers in bracket represent the number of samples in each category)

| Feature | CASMEII | SAMM | SMIC |
|---|---|---|---|
| Number of Samples | 255 | 159 | 164 |
| Participants | 35 | 32 | 16 |
| Ethnicities | Chinese | Chinese and 12 more | Chinese |
| Facial Resolution | 280 × 340 | 400 × 400 | 640 × 480 |
| Categories | Happiness (25), Surprise (15), Anger (99) Disgust (26), Sadness (20), Fear (1) Others (69) | Happiness (24), Surprise (13), Anger (20) Disgust (8), Sadness (3), Fear (7) Others (84) | Positive (51), Negative (70), Surprise (43) |

CK+ [33] contains 593 video clips collected from 123 subjects, among which 327 clips were labelled with AUs on the last frame of each. 7 emotion types are included: anger, contempt, disgust, fear, happiness, sadness and surprise. Each clip starts from a normal facial expression frame and ends at the apex frame. For our experiment, the last three frames from each corresponding video clip (belonging to the 5 emotion types we used) are selected while 852 images are employed.

Oulu-CASIA NIR&VIS facial expression database [34] contains the video from 80 subjects. Six typical expressions are collected, i.e. happiness, sadness, surprise, anger, fear and disgust. Two imaging systems, NIR (Near Infrared) and VIS (Visible Light) systems, were equipped for the data capturing. Furthermore, under each image system, three different illumination conditions were applied, i.e. normal indoor illumination, weak illumination (with a computer display on) and dark illumination (no visible lights). For our experiment, the last three frames of each video captured by VIS system under normal indoor illumination are extracted which provide 1200 images.

Jaffe [35] contains 219 images of 7 emotions displayed by 10 Japanese females. Each subject acted 7 types of expression, namely sadness, happiness, anger, disgust, surprise, fear and neutral. For our experiment, we selected images correspond to 5 emotion types to classify in micro-expression tasks, and finally extracted 151 images.

The MUGFE [36] includes 1032 video clips from 86 subjects. The database has two parts, one with six acted expressions and another with stimuli induced expressions. Each clip has 50 to 160 frames starting and ending at neutral expression with apex frames corresponding to respective emotion types within. Around the apex frames, we select 6 to 10 frames and finally have 8228 images for our experiment.

In total, 10431 images from four macro-expression databases are selected to pretrain the model. For each image, the facial area is segmented with AAM [40] and normalized to 224 × 224 pixels. During the pretraining, 1/10 of the images is used for validation while the rest for training. Additionally, three data augmentation methods are used with a selecting probability of 0.5, namely the color shift with maximum value of 20, rotation with maximum degree of 10 and smoothing with maximum window size of 6.

## 4.2 Model Implementation and Validation Methods

For the proposed micro-attention method, after initialization with ImageNet [38], at the stage of pretraining, we used batch gradient descent with a momentum of 0.9, the batch size is set to 50 and learning rate is initialized with 0.01 which would decrease 10 times smaller after every 20 epochs. The proposed model is implemented with Caffe [42]. In pre-training, the average recognition rate on 5 macro-expression types of the residual network (without attention units) is 99.33%. The same network with micro-attention would be applied during fine-tuning and testing.

At the stage of fine-tuning and testing, three validation methods are used: Holdout-database Evaluation (HDE) and Composite-database Evaluation (CDE) that were used in the MEGC 2018 [41], and traditional leave-one-subject-out validation (LOSO).

HDE and CDE are cross-database validations, where the CASME II and SAMM are used together. In HDE, one database is used for training and another for testing, while CDE is a Leave-One-Subject-Out validation (LOSO) with the two databases pooled together. For the first fold in HDE where CASME II is used as training set and SAMM is used as testing set, images in each micro-expression type is resampled according to the one having largest size w.r.t. the imbalanced dataset problem. Color shift (maximum value of 20), rotation (maximum degree of 8) are applied to augment the data with selecting probability of 0.5. For our method and other deep learning based approaches, the batch size is set to 10 and initial learning rate is set to 1e4 (momentum=0.9, weight decay=3e2) which is decreased 10 times smaller after every 10 epochs. For the second fold in HDE where SAMM is used as training set and CASME II is used as testing set, the resampling method, data augmentation and hyperparameter settings are the same. Metrics utilized here are Weighted Average Recall (WAR) and Unweighted Average Recall (UAR). The computation of WAR and UAR is shown below

$$WAR = \frac{\sum_{c=1}^{C} TP_c}{\sum_{c=1}^{C} N_c}, UAR = \frac{1}{C}\sum_{c=1}^{C} \frac{TP_c}{N_c} \qquad (3)$$

where $C$ is the number of categories, $TP_c$ is the number of true-positive samples and $N_c$ is the total number of samples under category $c$.

In CDE, 20 training-testing processes are prepared for SAMM and 26 for CASME II. For images in different expression types in the training set, the resampling method is also used to balance the classes. Data augmentation is also considered by cropping the four corners of the original image with size of $240 \times 240$ into a 'new' one with size of $224 \times 224$ and resizing it to the previous size. For our method and other deep learning based approaches, the batch size is set to 8 and initial learning rate is set to 1e3 (momentum=0.9, weight decay=5e6) which is decreased 10 times smaller after every 10 epochs. Metrics utilized here are average accuracy and F1-score.

The traditional LOSO is also included as to maintain the completeness of this work. Here, the three databases are used to test the model separately. The resampling method and the data augmentation methods mentioned above are together used for every training set. For our method and other deep learning based approaches, the batch size is set to 10 and initial learning rate is set to 1e3 (momentum=0.9, weight decay=5e4) which is decreased 10 times smaller after every 10 epochs. Metric used here are the average accuracy and F1 score.

Since the dataset settings in HDE and CDE are the same with what used in MEGC 2018, we directly used the baseline results achieved with LBP-TOP, HOOF and HOG3D [39], the results from study [32] as well as the state-of-the-art one [17] that won the challenge. For the detailed parameter settings, please refer to the original papers. To further demonstrate the advantage of parameter-saving in micro-attention, we compare it with another attention-based method [29] that designed for micro-gesture recognition with a more complex attention mechanism. The same transfer learning procedure is applied for all the deep learning based methods [32] [17] [29]. In traditional LOSO on the three databases, in order to perform a fair comparison, we only compare our method with the two deep learning based approaches [17] [29] because the other methods [39][32] are either not originally tested with the three databases separately or not using the traditional LOSO.

## 4.3 Experimental Results and Discussions

The experimental results plus the comparison of parameter sizes are provided in this subsection according to different validation methods. Discussions on these results are also given along with.

*A. Holdout-database Evaluation (HDE)*

Taking CASME II as the training set and SAMM as the test set, the residual network with proposed micro-attention achieved WAR of 0.559 and UAR of 0.427. With SAMM as the test set and CASME II as the training set, the residual network with proposed micro-attention achieved WAR of 0.584 and 0.341. The average WAR and UAR achieved by baseline methods [39], previous studies [17] [29] [32] and our method in two evaluation folds are summarized in Table 2. The dataset listed in each column is used as test set while another is used for training.

Table 2 | Average WAR and UAR achieved by different methods in HDE.

| Methods | WAR | | UAR | |
|---|---|---|---|---|
| | SAMM | CASMEII | SAMM | CASMEII |
| LBP-TOP [39] | 0.338 | 0.232 | 0.327 | 0.316 |
| HOOF [39] | 0.441 | 0.265 | 0.349 | 0.346 |
| 3D-HOG [39] | 0.353 | 0.373 | 0.269 | 0.187 |
| ELRCN [32] | 0.485 | 0.384 | 0.382 | 0.322 |
| Residual Network [17] | 0.544 | 0.578 | **0.440** | 0.337 |
| Residual Network with complex attention [29] | 0.529 | 0.573 | 0.393 | 0.319 |
| **Residual Network with Micro-Attention** | **0.559** | **0.584** | 0.427 | **0.341** |

We can see from Table 2 that the proposed method yielded better WAR (0.559 and 0.584) in inter-database validations and better UAR (0.341) when tested on CASMEII. Specially, the proposed attention units rendered the residual network better results comparing with the previous study [17] that did not use attention mechanism. However, the UAR (0.427) achieved with our method on SAMM is slightly lower than [17] (0.440), while the WAR achieved is better. Although the resampling method was conducted, given the computation of UAR which is considering the distribution of categories, such results are probably due to the imbalanced training set (CASMEII) that categories with larger size would be easier to catch the attention of the model. Another HDE experiment with Residual Network [17] on the two datasets excluding the transfer learning phase was conducted, but we found that the model failed to converge so the results are not reported. Such situation is normally seen when directly applying deep networks on small datasets.

On the other hand, the parameter size and training time per iteration of the three residual based methods are reported in Table 3. Given that micro-attention is designed to utilize the existing self-learned feature maps which avoid creating extra computational layers so reduced the size of parameters, the proposed method still achieves better results than previous attention-based method [29]. As the attention mechanism introduced in [29] is bringing in notable parameters while the training data is comparatively small, probably due to overfitting, the WAR and UAR achieved by which are even lower than normal residual network without attention [17].

Table 3 | Parameter size and training time per iteration of three residual network methods.

| Methods | Parameter size (million) | Time per iteration (second) |
|---|---|---|
| Residual Network [17] | 4.9 | 0.95 |
| Residual Network with complex attention [29] | 11.2 | 1.5 |
| **Residual Network with Micro-Attention** | **5.9** | **1.1** |

## *B. Composite Database Evaluation (CDE)*

For the CDE on SAMM and CASME II databases, the average accuracy and F1-score of each method are summarized in Table 4. The parameter size of each network-based method remains unchanged.

Table 4 | Average recognition accuracies and F1-scores of different methods on SAMM and CASME II in CDE.

| Methods | Accuracy | F1-score |
|---|---|---|
| LBP-TOP | 0.524 | 0.400 |
| HOOF | 0.527 | 0.404 |
| 3D-HOG | 0.436 | 0.271 |
| ELRCN [32] | 0.570 | 0.411 |
| Merghani et al [39] | 0.718 | 0.579 |
| Residual Network [17] | 0.747 | 0.640 |
| Residual Network with complex attention [29] | 0.625 | 0.489 |
| **Residual Network with Micro-Attention** | **0.763** | **0.668** |

As shown, the proposed model has better generalization ability to unseen subjects. We owe this to the design of micro-attention units because the subject identity, which can be deemed as noise to the micro-expression recognition, is out of the attention. Finally, the confusion matrix for the classification result in CDE of our proposed method can be seen in Table 5. Aside from better results the model achieved in general, the classification on happiness and anger is much better than other categories. This may due to the different biomechanical natures of the two emotions which both have strong facial movement intensities. As is also shown in the confusion matrix, the classification of disgust and sadness is under chance level and most of the two emotions are classified as anger. This could be that disgust and sadness share many similarities with anger w.r.t. the types of AU characterizing them. On the other hand, another significant reason lays in the unbalanced distribution of samples, the number of samples labelled with Disgust or Sadness is much smaller than other types. Consequently, we can see that a better future of micro-expression relies on the development of modeling as well as bigger micro-expression database with better category distribution.

Table 5 | Confusion matrix of the result yielded by proposed method in CDE.

|           | Happiness | Surprise | Anger | Disgust | Sadness |
|-----------|-----------|----------|-------|---------|---------|
| Happiness | **81.63** | 12.24    | 6.12  | 0.00    | 0.00    |
| Surprise  | 17.86     | **53.57**| 10.71 | 7.14    | 10.71   |
| Anger     | 2.52      | 0.00     | **94.12** | 1.68 | 1.68    |
| Disgust   | 17.65     | 0.00     | 35.29 | **47.06**| 0.00   |
| Sadness   | 4.35      | 8.70     | 43.47 | 0.00    | **43.48**|

## C. Traditional Leave One Subject Out Validation (LOSO)

For the LOSO on SAMM, CASME II and SMIC databases separately, results are reported in Table 6. Generally, results achieved in traditional LOSO is better than what reported in HDE but is lower than results in CDE. For the former situation, it is reasonable because HDE is a strict validation method on testing the generalization ability of a model across different datasets. For the later one, the accuracies achieved in CDE are higher, which is probably because that the respective training set is bigger as the two datasets (SAMM and CASME II) are combined. Still, our method yields the best performances on the three databases.

Table 6 | Average recognition accuracy and F1-score of different methods on SAMM, CASME II and SMIC in traditional LOSO.

| Methods | SAMM | | CASME II | | SMIC | |
|---------|------|------|----------|------|------|------|
|         | Accuracy | F1-score | Accuracy | F1-score | Accuracy | F1-score |
| Residual Network [17] | 0.456 | 0.383 | 0.622 | 0.464 | 0.415 | 0.406 |
| Residual Network with complex attention [29] | 0.471 | 0.306 | 0.627 | 0.473 | 0.494 | 0.448 |
| **Residual Network with Micro-Attention** | **0.485** | **0.402** | **0.659** | **0.539** | **0.494** | **0.496** |

### 4.4   Impact analysis through feature visualization

Results in HDE, CDE and LOSO have demonstrated the better recognition ability of our approach comparing with state-of-the-art ones [17] [29] [32] [39]. In this subsection, we make efforts to interpret the modeling behavior of our method through feature visualization, which is usually employed to explain the decision of a model given its focus paid to the input data.

From the residual network [17] without attention units, we first extract the feature map from the last residual block with forward propagation, using inputs labelled with disgust, sadness and anger respectively. Similarly, such feature maps are also extracted from the last micro-attention unit of our network and from the last attention unit of the network proposed in [39]. The feature visualizations are given in Figure 4.

At lower-level, the edge information (representing the face) is mainly learned while the information contributing to the classification is learned at higher-level. We can see from the figure that the differences of focused areas (marked with higher attention intensity) of the three networks lead to different classification results, where our method with micro-attention units produced the correct classification labels and the other two failed for the recognition of one or all emotion types. To better illustrate the impact of the proposed micro-attention units, the distributions of facial AUs [38] related to the input emotions of disgust, sadness and anger are demonstrated in Figure 5. Detailed AU combinations for those emotions are summarized in Table 7.

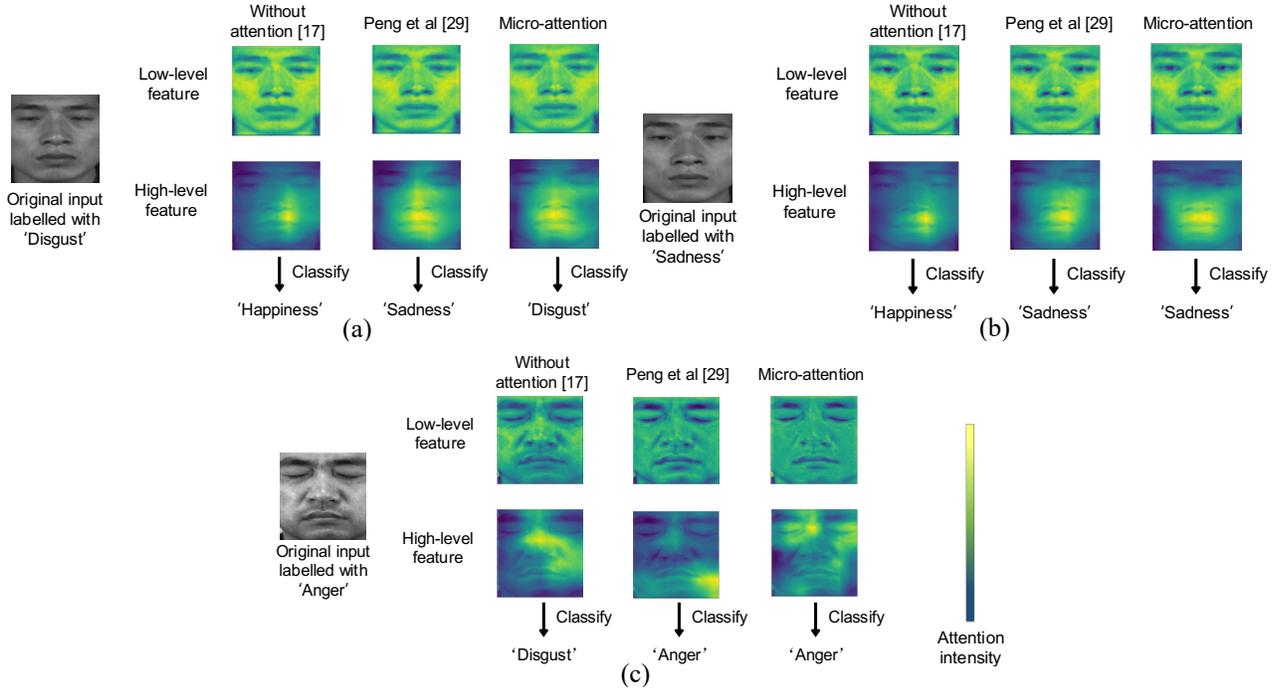

Figure 4. Feature map visualizations of network without attention, with more complex attention proposed in [29] and with micro-attention on the recognition of: (a) disgust; (b) sadness; (c) anger.

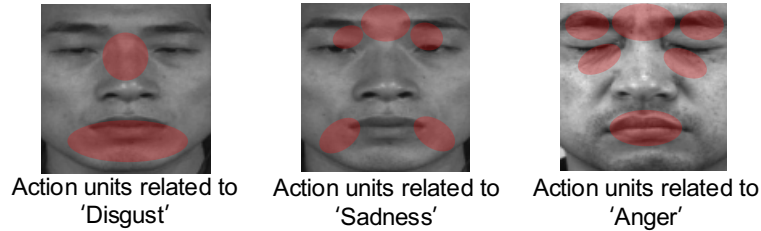

Figure 5. Distributions of AUs related to disgust, sadness and anger.

Table 7 | Emotion types and the corresponding AU combinations.

| Emotion type | Action unit combinations |
|---|---|
| Disgust | AU9 (nose wrinkler), AU15 (lip corner depressor), AU16 (lower lip depressor) |
| Sadness | AU1 (Inner brow raiser), AU4 (brow lower), AU15 (lip corner depressor) |
| Anger | AU4 (brow lower), AU5 (upper lid raiser), AU7 (lid tightener), AU23 (lip tightener) |

For the recognition of disgust, with the attention unit proposed in [39] and micro-attention units, the two models started to pay attention to the nasal and lip areas that are correlated with the genuine expression. Without attention mechanism, the original residual network [17] produced the wrong decision with the focus on less relevant areas (canthus) under the same training process. Such same situation also happened for the recognition of sadness. More obvious pattern is found in the recognition of anger, where the distribution of attention-paid areas of our method better covered the relevant AUs and yielded the correct result. Although the previous method [39] also produced the right classification, it is more likely to be an inference at chance-level because the focused area (platysma) are much less relevant with anger. For the slightly overlapped AU distributions around nasal and mouth areas of anger and disgust, the residual network without attention mechanism [17] misclassified anger as disgust because of its focused areas which mostly covered the AUs representing the latter one.

Generally speaking, the effective role of attention mechanism in macro- and micro-expression recognition is demonstrated through the experiments. However, it is also learned from the comparison results that, for the design of specific attention unit, one thing should be noticed is the impact of extra parameters. Especially for the recognition task on small dataset, such factor would particularly hinder the performance of using attention mechanism.

# 5 Conclusion

In this work, we proposed a novel attention mechanism called micro-attention to help the network to focus on facial areas of interest for the recognition of micro-expression. The mechanism is implemented as micro-attention units to be used with the block-wise design of residual network, so as to take the advantage of self-learned multi-scale spatial features within each residual block. To alleviate the overfitting risk of training deep network on small dataset, such attention unit is further designed to reduce the use of extra computational layers so as to avoid increasing notable parameters, while a transfer learning strategy is also employed using ImageNet and four macro-expression databases. Specifically, during the transfer learning process, the network is first initialized on ImageNet and then pre-trained on four popular macro-expression databases. After fine-tuned on three benchmark databases (CASMEII, SAMM and SMIC), the proposed method achieved better recognition accuracy than several state-of-the-art methods. Specially, comparing with two residual based methods, the visualized high-level feature map demonstrates the effectiveness of the proposed attention units on capturing the facial areas which better correlated to the underlying emotion.

# 6 Conflict of Interest

The authors share no conflicts toward this work.

# 7 Author Contributions

Chongyang Wang and Min Peng designed the theory and experiments. Tao Bi and Tong Chen provided suggestions during experiments. All authors contributed to the writing.

# 8 Funding

Chongyang Wang is supported by UCL Overseas Research Scholarship (ORS) and UCL Graduate Research Scholarship (GRS).

# 9 Acknowledgement

Thanks to Professor Nadia Berthouze at University College London who provided abundant suggestions during writing and submitting this paper.